\documentclass[runningheads]{llncs}

\usepackage[T1]{fontenc}
\usepackage{graphicx}
\usepackage[hidelinks]{hyperref}
\usepackage{booktabs} 
\usepackage{multirow}
\usepackage{tabularx}
\usepackage{amsmath}
\usepackage{amsfonts}
\usepackage{tikz}
\usepackage{algorithm}
\usepackage{algpseudocode}
\usepackage{listings}
\usepackage{inconsolata}
\usepackage{microtype}  
\usepackage{placeins}
\usepackage{float} 

\usepackage{tcolorbox}

\newsavebox\CBox
\newcommand*\textBF[1]{\sbox\CBox{#1}\resizebox{\wd\CBox}{\ht\CBox}{\textbf{#1}}}

\lstdefinestyle{promptstyle}{
  basicstyle=\ttfamily\small,
  columns=fullflexible,
  breaklines=true,
  breakatwhitespace=false,
  frame=single,
  showstringspaces=false,
  keepspaces=true,
  tabsize=2
}

\usetikzlibrary{positioning, shapes.geometric, arrows.meta}

\begin{document}

\titlerunning{SERC: LDPC-Inspired Semantic Error Correction}

\title{SERC: LDPC-Inspired Semantic Error Correction for Retrieval-Augmented Generation}



\makeatletter
\renewcommand{\thefootnote}{\fnsymbol{footnote}}
\renewcommand*\@fnsymbol[1]{\ifcase #1 \or * \or † \else \@ctrerr \fi}
\makeatother

\author{
Gyumin Kim\inst{1}\thanks{These authors contributed equally to this work.} \orcidID{0009-0003-9422-1735} \and
Juhwan Park\inst{2}\protect\footnotemark[1] \orcidID{0009-0005-2428-6124} \and
Jaeha Kim\inst{1}\protect\footnotemark[1] \orcidID{0009-0002-2115-5899} \and
Seunggyun Han\inst{3} \orcidID{0009-0000-4670-1695} \and
Kyungrak Son\inst{1}\protect\footnotemark[2] \orcidID{0000-0002-2983-985X} \and
Ikbeom Jang\inst{2}\thanks{Corresponding authors.} \orcidID{0000-0002-6901-983X} 
}

\authorrunning{G. Kim et al.}

\institute{
    Department of Information Communications Engineering, Hankuk University of Foreign Studies, Republic of Korea \and
    Division of Computer Engineering, Hankuk University of Foreign Studies, Republic of Korea \and
    Department of Statistics, Hankuk University of Foreign Studies, Republic of Korea\\
    \email{\{rhzs1208, amry0719, jaehakim, mashan120, krson, ijang\}@hufs.ac.kr}
}
\maketitle              
\begin{abstract}
While Large Language Models (LLMs) have demonstrated remarkable capabilities, their reliability is significantly compromised by hallucinations. Existing intrinsic self-correction methods attempt to address this, but often fail due to self-bias, where models struggle to identify errors in their own outputs without external verification. To overcome these limitations, we propose the LDPC-inspired semantic error correction for retrieval-augmented generation (SERC), providing a theoretical framework to interpret and mitigate LLM hallucinations. We reformulate the text generation process as a semantic noisy channel, treating generated responses as noise-corrupted codewords. Inspired by low-density parity-check (LDPC) codes, SERC employs a sparse verification strategy: instead of exhaustively checking all facts, it generates low-density verification queries and validates them against external evidence to efficiently detect and correct errors. We evaluate SERC on LongForm Bio and TruthfulQA benchmarks using Llama-3-8B and Qwen2.5-14B. Experimental results demonstrate that SERC outperforms both intrinsic self-correction methods and strong retrieval-augmented baselines, demonstrating significant gains especially in factual precision (FactScore). Notably, SERC enables small language models (SLMs) to surpass the performance of larger baselines in hallucination reduction and information preservation. Our findings demonstrate that SERC provides a training-free, model-agnostic solution that significantly reduces verification overhead compared to dense methods, achieving an optimal trade-off between cost and fidelity in resource-constrained environments. The code and data are available at \url{https://github.com/labhai/SERC}.

\keywords{Large Language Models \and Hallucination Mitigation \and Error Correcting Codes \and Retrieval-Augmented Generation \and Semantic Noisy Channel.}
\end{abstract}
%
%

\section{Introduction}

Recent advancements in large language models (LLMs) \cite{brown2020language} are constrained by hallucinations---content inconsistent with factual reality \cite{ji2023survey}. This structural limitation poses significant risks in high-stakes domains like healthcare and law \cite{weidinger2021ethical}, making the detection and correction of reasoning errors a critical challenge.

Existing self-correction methods, such as Chain-of-Verification (CoVe) \cite{dhuliawala2023cove}, largely rely on intrinsic reasoning but suffer from Self-Bias, where initial biases propagate into the verification phase \cite{huang2024large}. Implementing such procedures in small language models (SLMs) is further complicated by their limited reasoning capacity \cite{wei2022emergent}. We argue that incorporating Retrieval-augmented generation (RAG) is a prerequisite for effective self-correction, especially in resource-constrained SLM environments \cite{lewis2020retrieval}.

To address these limitations, we redefine hallucinations through an information-theoretic lens, interpreting LLM outputs as signals transmitted via an imperfect \textit{semantic channel} \cite{qin2021semantic}. By reformulating hallucination as a probabilistic inference problem where generated text is a noisy observation of unobserved factual latent variables \cite{kingma2013auto}, we apply classical error-correction principles to recover the ground-truth manifold.

Building upon this, we propose the SERC (Semantic Error-Reduction and Correction) framework. Inspired by the design philosophy of low-density parity-check (LDPC) codes \cite{gallager1962ldpc}, SERC establishes a \textit{low-density verification plan} to efficiently expose error patterns. Detected errors are rectified through a procedure analogous to the belief propagation (BP) algorithm \cite{pearl1988probabilistic}, propagating updated beliefs to restore global semantic coherence. Our primary contributions are threefold: (1) we propose a semantic channel model that interprets LLM hallucinations through the lens of error correcting codes (ECC), providing a theoretical basis for self-correction; (2) we implement the SERC framework, which utilizes low-density verification on atomic facts to efficiently detect errors without additional training; and (3) we demonstrate through experiments on LongForm Bio and TruthfulQA that SERC significantly outperforms baselines like CoVe, specifically enhancing the reliability of SLMs.


\section{Related Works}

\textbf{LLM Correction and Retrieval-Augmented Generation: }
LLM hallucinations stem from training data tails and parametric gaps \cite{ji2023survey,mallen2023trust}. Intrinsic self-correction (e.g., CoVe \cite{dhuliawala2023cove}) often succumbs to \textit{self-bias} without external grounding \cite{huang2024large}. RAG \cite{lewis2020retrieval} mitigates this, but adaptive variants (Self-RAG \cite{asai2024selfrag}) require costly fine-tuning and struggle with noisy retrieval \cite{yoran2024robust}. Advanced systems optimizing retrieval pipelines (CRAG \cite{yan2024corrective}, MIGRES \cite{zhang2024migres}, Adaptive RAG \cite{jeong2024adaptive}) focus primarily on information gathering, remaining vulnerable to semantic noise introduced during generation. Conversely, SERC transcends mere retrieval, functioning as an information-theoretic error correction (ECC) mechanism to systematically rectify corrupted semantics. Operating \textit{training-free} and \textit{model-agnostic} at the \textit{semantic proposition level}, SERC's \textit{low-density} verification provides superior cost-efficiency over exhaustive baselines like RARR \cite{gao2023rarr}.

\textbf{Information Theory and Coding: }
Error correcting codes (ECC) \cite{shannon1948mathematical} ensure reliable transmission. Notably, low-density parity-check (LDPC) codes \cite{gallager1962ldpc} enable efficient decoding via sparse parity-check structures, represented as Tanner Graphs \cite{tanner1981recursive} and decoded via iterative belief propagation (BP) \cite{pearl1988probabilistic}. In semantic communication, systems like DeepSC \cite{xie2021deep} transmit semantic meaning rather than mere bits, preserving semantics over noisy channels. SERC adapts these principles for hallucination mitigation, interpreting LLM outputs as noise-corrupted codewords and verifications as sparse parity checks. The resulting Tanner-style graph links factual propositions with evidence, enabling structured hallucination correction from an information-theoretic perspective.


\section{Information Theoretic Abstraction}
\label{sec:abstraction}

\subsection{Semantic Channel Modeling: LLM as a Noisy Channel}
\begin{figure}[]
    \vspace{-4mm}
    \centering
    \includegraphics[width=\linewidth, trim=0 180 0 180, clip]{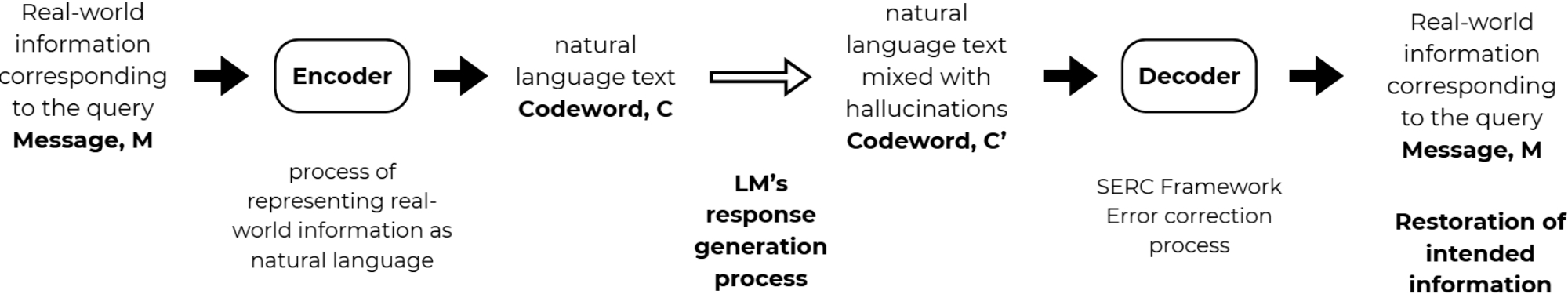}

    \caption{The proposed Semantic Channel Model. The generation process of an LLM is modeled as a noisy channel where hallucinations are treated as semantic noise, and SERC acts as the decoder to restore original information.}
    \label{fig:system_model}
\end{figure}
To address the self-bias in intrinsic self-correction \cite{dhuliawala2023cove,huang2024large}, we model the LLM generation process as a semantic noisy channel. By drawing a formal analogy to classical information theory \cite{shannon1948mathematical,qin2021semantic}, we decompose the QA process into five core components as illustrated in Fig.~\ref{fig:system_model}:

\noindent\textbf{Source:} The origin of information, representing the real-world entity or subject of inquiry that the user intends to learn about.

\noindent\textbf{Message ($M$):} The set of objective ground-truth facts existing in the real world regarding the subject. $M$ defines the complete space of factual validity independent of its linguistic expression.

\noindent\textbf{Codeword ($C$):} The ideal, hallucination-free natural language response constructed solely using facts belonging to message $M$. For example, for a query regarding Einstein, a potential \textit{Codeword} is: ``Einstein was born in Germany and published the theory of relativity.''

\noindent\textbf{Semantic Noisy Channel:} The stochastic LLM generation process. While an ideal model outputs $C$, actual LLMs act as a channel injecting semantic noise \cite{ji2023survey}, distorting the ideal codeword into a corrupted observation $C' = C \oplus \text{Noise}$, where $\oplus$ denotes the superposition of semantic distortions(see Supp. Sec. 5).

\noindent\textbf{Decoder (SERC):} The proposed framework that serves as a semantic decoder. Analogous to LDPC \cite{gallager1962ldpc} and BP \cite{pearl1988probabilistic}, SERC reconstructs the original codeword $C$ from the noisy observation $C'$ to restore factual fidelity to the original message $M$.

\subsection{Manifold of Truth and Operational Approximation}
\label{sec:manifold}

Building on the semantic channel abstraction, SERC operates as an external error correction layer that rectifies the noisy output $C'$ without accessing the LLM's latent space. We define the Operational Fact Set $F = \{f_{k,i} \mid 1 \le k \le n, 1 \le i \le m_k\}$ as the collection of atomic factual propositions extracted from $C'$. Here, $f_{k,i}$ denotes the $i$-th atomic fact derived from the $k$-th sentence $s_k$, serving as the individual variable nodes for our decoding graph.

The objective of SERC is to project the corrupted set $F$ onto the manifold of truth ($\mathcal{M}_{\text{truth}}(Q)$), defined as the space of all fact sets derived from ideal, hallucination-free responses $\mathcal{C}^*(Q)$:
\begin{equation}
    \mathcal{M}_{\text{truth}}(Q) = \{ \text{Facts}(C^*) \mid C^* \in \mathcal{C}^*(Q) \}
\end{equation}
Since $\mathcal{M}_{\text{truth}}(Q)$ is unobservable, SERC operationally approximates it using the subspace of facts consistent with external evidence retrieved via RAG.

To perform this projection efficiently, we adopt a graph-based strategy inspired by low-density parity-check (LDPC) codes. Instead of dense, exhaustive verification, we construct a sparse Tanner graph where multiple facts (variable nodes) are verified by a single, grouped verification query (check node). This sparsity minimizes the computational overhead of verification (LLM/RAG calls). Finally, the correction process mimics belief propagation (BP); a local correction in one fact (e.g., entity type) logically propagates to related sentences, ensuring the reconstructed text converges to global semantic consistency.

\section{Proposed Methodology: SERC Framework}
\label{sec:methodology}

We propose the SERC (Semantic Error-Reduction and Correction) framework. Building on the theoretical foundation established in Section~\ref{sec:abstraction}, SERC instantiates the semantic decoder. We map the abstract components of the channel model to concrete RAG operations: the noisy codeword $C'$ is instantiated as the initial LLM response $R_{\text{init}}$, and the parity check constraints are implemented via sparse verification queries. As illustrated in Algorithm~\ref{alg:serc_compact}, the framework operates sequentially through three mathematically defined phases to align the response with the Manifold of Truth.


\subsection{Phase 1: Coarse Alignment and Entity Firewall}
The process initiates with the generation of an initial response $R_{\text{init}}$ (corresponding to the noisy observation $C'$) from the language model $LM$ given a user query $Q$:
\begin{equation}
    R_{\text{init}} \sim P_{LM}(y \mid Q)
\end{equation}
Standard RAG approaches often fail when the initial generation suffers from source confusion (e.g., confusing two people with the same name). In channel coding terms, this represents a Synchronization Error, where the decoder attempts to decode a signal using the wrong codebook. To mitigate this, we introduce the entity firewall mechanism.
Let $\mathcal{T}(\cdot)$ be a topic entity extraction function. We extract the core subject entity from both the model's internal knowledge ($T_{\text{model}} = \mathcal{T}(R_{\text{init}})$) and external evidence retrieved via a search module $\mathcal{R}$ ($T_{\text{rag}} = \mathcal{T}(\mathcal{R}(Q))$).
The firewall verifies the consistency between these two entities(see Supp. 1.1 for the exact judge prompt):
\begin{equation}
    R_{\text{init}} = 
    \begin{cases} 
    LM(Q, \mathcal{R}(Q)) & \text{if } \text{Consistency}(T_{\text{model}}, T_{\text{rag}}) = \text{False} \quad (\text{Hard Reset}) \\
    R_{\text{init}} & \text{otherwise}
    \end{cases}
\end{equation}
If a mismatch is detected, a \textit{Hard Reset} is triggered, forcing the model to regenerate the baseline using the retrieved context to align the semantic trajectory before fine-grained verification.

\subsection{Phase 2: Fact Decomposition and Sparse Verification}

\begin{figure}[t]
    \centering
    \includegraphics[width=0.90\linewidth, trim=0 35mm 0 15mm, clip]{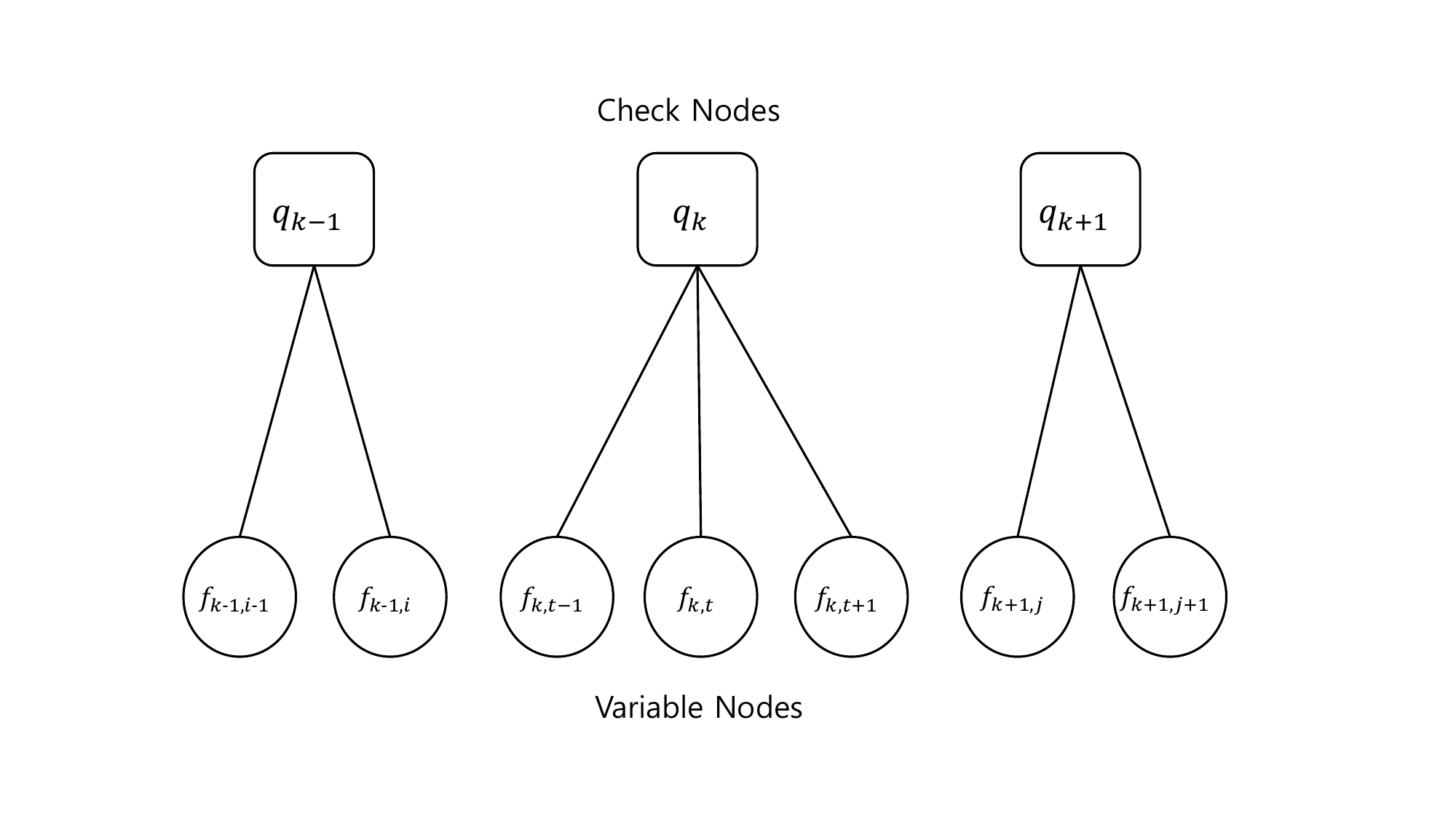}
    \caption{\textbf{Semantic Tanner Graph Structure.} The bottom nodes (Variable Nodes) represent atomic facts extracted from sentences, while the top nodes (Check Nodes) represent grouped verification queries. This bipartite structure enables efficient sparse verification.}
    \label{fig:tanner_graph}
\end{figure}

To perform granular error correction,we first decompose the continuous signal $R_{\text{init}}$ into discrete semantic symbols. The baseline response $R_{\text{init}}$ is decomposed into a set of sentences $S = \{s_1, \dots, s_n\}$. For each sentence $s_k$, we extract a subset of atomic facts $F_k = \{f_{k,1}, \dots, f_{k,m_k}\}$ using a dedicated prompt(see Supp. 1.2).

\noindent\textbf{Tanner Graph Construction}
We map the verification problem to a Semantic Tanner Graph $G=(V, N_c, \mathcal{A})$, analogous to low-density parity-check (LDPC) codes, where $\mathcal{A}$ represents the set of arcs connecting facts to verification queries, as illustrated in Figure~\ref{fig:tanner_graph}:

\noindent\textbf{Variable Nodes ($V$):} Represent individual atomic facts $f_{k,i}$ (bottom nodes).

\noindent\textbf{Check Nodes ($N_c$):} Represent verification queries (top nodes). To optimize computational cost, we adopt a sparse verification strategy. Instead of verifying each fact independently, we generate a single comprehensive query $q_k = \text{GenQ}(F_k)$ for each sentence group $F_k$, which functions as the Check Node constraint.(details in Supp. 1.3)

\noindent\textbf{Syndrome Detection}
Using the query $q_k$, we retrieve external context via $\mathcal{R}$ and generate a concise evidence summary $E_k$ using the backbone model. The validity of each fact is then evaluated by a verifier $\mathcal{V}$ (refer to Supp. 1.3 for the verdict prompt) to compute the Semantic Syndrome $\sigma_{k,i}$:
\begin{equation}
    \sigma_{k,i} = \mathcal{V}(f_{k,i}, E_k) \in \{ \text{SUP}, \text{CON}, \text{NF} \}
\end{equation}
where $\text{SUP}$ (Supported) indicates a valid fact, $\text{CON}$ (Contradicted) indicates a hallucination, and $\text{NF}$ (Not Found) indicates unverifiable information.
Contradicted facts are collected in a syndrome buffer $\mathcal{B}_{\text{syn}}$, while unverifiable facts are assigned to a deletion set $\mathcal{S}_{\text{del}}$ for pruning.

\subsection{Phase 3: Logic Propagation via BP-inspired Heuristic}
The final phase corrects detected errors and reconstructs the text. We define the correction process as a function $\text{BP}(\mathcal{B}_{\text{syn}})$, which outputs a Correction Map $\Phi$. This map assigns a corrected fact $f'_{k,i}$ to each $f_{k,i}$ associated with a non-zero syndrome. 

While traditional BP algorithms rely on iterative message passing until convergence, such recursion is computationally prohibitive in the semantic domain due to the high latency of LLM inference. To address this, we propose a \textit{Single-Pass Unfolding} of the belief propagation algorithm. We propagate the ``beliefs'' of verified facts (acting as clamped variable nodes) exactly once through the semantic dependency graph. This acts as a \textit{first-order approximation} of BP, capturing the core mechanism of message passing---updating dependent variables based on reliable evidence---while maintaining inference efficiency. Crucially, we empirically observed that semantic dependencies in hallucinations are predominantly local (1-hop); thus, a single-pass approximation provides sufficient error-correction capability without the need for iterative convergence.

\noindent\textbf{Step 1: Local Belief Update}
Based on the syndrome $\sigma_{k,i}$, we update the belief of each variable node (fact) to produce a corrected fact $f'_{k,i}$:
\begin{equation}
    f'_{k,i} = 
    \begin{cases} 
    \text{Correct}(f_{k,i}, E_k) & \text{if } \sigma_{k,i} = \text{CON} \\
    \emptyset \text{ (Prune)} & \text{if } \sigma_{k,i} = \text{NF} \\
    f_{k,i} & \text{if } \sigma_{k,i} = \text{SUP}
    \end{cases}
\end{equation}
Crucially, we employ a \textit{Logic Propagation} prompt. If a fundamental attribute is corrected, this change propagates to logically dependent facts within the same group using the shared evidence $E_k$.

\noindent\textbf{Step 2: Sequential Reconstruction}
To ensure global coherence, the text is reconstructed sequentially. To prevent the regurgitation of hallucinations caused by anchoring bias, we adopt a \textit{Fact-to-Text} generation strategy, explicitly excluding the original sentence $S_k$. Consequently, the generation of the corrected sentence $s'_k$ is conditioned \textit{solely} on the corrected facts $F'_k$ and the history of previously reconstructed sentences $H_{k-1}$ (represented as $C_{\text{draft}}$ in Algorithm~\ref{alg:serc_compact}):
\begin{equation}
    s'_k = LM_{\text{rewrite}}(F'_k \mid H_{k-1})
\end{equation}
This auto-regressive reconstruction ensures that local corrections propagate through the narrative flow. Finally, a lightweight Polishing module smooths the reconstructed draft to yield the final response $C_{\text{final}}$ (see Supp. 1.4 for details).
\FloatBarrier
\begin{algorithm}[H]
\small
\caption{Compact SERC Execution Flow}
\label{alg:serc_compact}
\begin{algorithmic}[1]
\setlength{\itemsep}{0pt}
\setlength{\parskip}{0pt}

\Require Query $Q$, Model $LM$, Retriever $\mathcal{R}$
\Ensure Final Response $C_{\text{final}}$

\State \textbf{1. Alignment:} $R_{\text{init}} \gets LM(Q)$
\If{$\text{Consistency}(\mathcal{T}(R_{\text{init}}), \mathcal{T}(\mathcal{R}(Q))) = \text{False}$} 
    \State $R_{\text{init}} \gets LM(Q, \mathcal{R}(Q))$ \Comment{Firewall (Hard Reset)}
\EndIf

\State \textbf{2. Detection:} $S \gets \text{Split}(R_{\text{init}})$; $\mathcal{B}_{\text{syn}} \gets \emptyset$; $\mathcal{S}_{\text{del}} \gets \emptyset$
\For{each $S_k \in S$}
    \State $F_k \gets \text{Facts}(S_k)$; $E_k \gets LM(\text{GenQ}(F_k) \mid \mathcal{R}(\text{GenQ}(F_k)))$
    \For{$f \in F_k$}
        \State $\sigma \gets \mathcal{V}(f, E_k)$ \Comment{Calculate Syndrome}
        \State \textbf{if} $\sigma = \text{CON}$ \textbf{then} $\mathcal{B}_{\text{syn}}.\text{add}(f, E_k)$
        \State \textbf{if} $\sigma = \text{NF}$ \textbf{then} $\mathcal{S}_{\text{del}}.\text{add}(f)$ \Comment{Mark for deletion}
    \EndFor
\EndFor

\State \textbf{3. Correction:} $\Phi \gets \text{BP}(\mathcal{B}_{\text{syn}})$ \Comment{Compute Correction Map}
\State $\forall f \in \mathcal{S}_{\text{del}}, \Phi[f] \gets \emptyset$ \Comment{Map 'NF' to empty}

\State \textbf{4. Reconstruction:} $C_{\text{draft}} \gets \emptyset$
\For{each $S_k \in S$}
    \State $F'_k \gets \{ \Phi[f] \text{ if } f \in \text{dom}(\Phi) \text{ else } f \mid f \in F_k \}$
    \State $C_{\text{draft}}.\text{append}(LM_{\text{rewrite}}(F'_k \mid C_{\text{draft}}))$
\EndFor

\State \Return $\text{Polish}(Q, C_{\text{draft}})$

\Statex \hrulefill
\Statex \textbf{Definitions:}
\Statex $\mathcal{T}(\cdot)$: Topic Extractor \quad $\text{Consistency}(\cdot)$: Entity Consistency Checker
\Statex $\text{Facts}(\cdot)$: Fact Decomposition \quad $\text{GenQ}(\cdot)$: Verification Query Generator
\Statex $\mathcal{V}(\cdot)$: Fact Verifier \quad $\text{BP}(\cdot)$: Belief Propagation Correction
\Statex $\Phi$: Correction Map \quad $LM_{\text{rewrite}}(\cdot)$: Fact-to-Text Generator
\end{algorithmic}
\end{algorithm}
\FloatBarrier


\section{Experiment}
\label{sec:experiment}

\subsection{Overview and Model Selection}
We evaluate SERC’s effectiveness regarding factual precision, token efficiency, and scalability across model scales. We utilize Llama-3-8B-Instruct and Qwen2.5-14B-Instruct as backbone models. Llama-3-8B serves as a testbed for knowledge-deficient SLMs on consumer GPUs, while Qwen2.5-14B, with its expanded capacity and distinct training distribution, verifies the scalability of our semantic correction mechanism.

\subsection{Benchmarks and Evaluation Metrics}
Experiments use a generation temperature of 0 to ensure reproducibility.

\noindent\textbf{LongForm Bio \& FactScore}: We evaluate long-form factual consistency using \textbf{FActScore}~\cite{min2023factscore}, which decomposes text into atomic propositions and verifies them against a reliable knowledge base.

\noindent\textbf{Fact Preservation Rate}~\cite{min2023factscore}: To distinguish information refinement from defensive deletion, we measure the ratio of facts retained relative to the initial answer, ensuring SERC does not merely truncate uncertain content.

\noindent\textbf{TruthfulQA \& LLM-as-a-Judge}: We assess the model's tendency to mimic human falsehoods~\cite{lin2022truthfulqa}. We replace traditional selection-based metrics (MC1/MC2) with the \textbf{LLM-as-a-Judge} paradigm~\cite{zheng2024judging}. To mitigate bias, Gemini-3-Pro(gemini-3-pro-preview) and GPT-5.1 (gpt-5.1-2025-11-13) serve as dual evaluators to ensure objective scoring~\cite{zheng2024judging}(rubric and prompts are in Supp. Sec. 2).

\noindent\textbf{Cost Analysis}: Total token usage is analyzed and compared against exhaustive baselines like RARR~\cite{gao2023rarr} to demonstrate the efficiency of our low-density verification strategy.

\subsection{Baselines and Comparative Methods}
To verify the effectiveness of SERC, we compare it against the following baselines and state-of-the-art methodologies(prompts in Supp. Sec. 4):

\noindent\textbf{Initial Answer:} The raw, uncorrected response generated by the backbone model (Llama-3-8B or Qwen2.5-14B).

\noindent\textbf{CoVe(Chain-of-Verification)\cite{dhuliawala2023cove}:} An intrinsic self-correction method that uses a multi-step reasoning process (generating verification questions and answering them) without external tools.

\noindent\textbf{CoVe+RAG:} An augmented version of CoVe provided with the same external knowledge as SERC, used to isolate the performance gains attributable to our ECC-inspired framework versus simple retrieval.

\noindent\textbf{RARR(Research and Revise)\cite{gao2023rarr}:} A retrieval-augmented baseline that exhaustively researches and revises each claim in the generated text using external evidence.

\noindent\textbf{Re-Ex(Revising after Explanation)\cite{kim2024reex}:} A baseline focused on the iterative re-extraction and verification of claims from retrieved contexts to ensure maximum factual grounding.

\subsection{Implementation Details}
Experiments were conducted on a single NVIDIA A100 (80GB) GPU using the \texttt{transformers} library. Backbone models were loaded with 16-bit (bfloat16) quantization to optimize memory usage. For retrieval, we utilized the \textBF{Tavily Search API} with an ``advanced'' search depth and a top-$k$ of 8, truncating retrieved contexts at 20,000 characters to prevent overflow. 

The pipeline utilized deterministic decoding ($T=0.0$) for most phases, including fact extraction, verification, and reconstruction, to ensure factual consistency. Temperature $T=0.1$ was applied exclusively during the final polishing stage to enhance linguistic fluency, with \texttt{max\_new\_tokens=512} for all generative tasks. Implementation details are described in the supplement and our GitHub repo.---e.g., prompt templates, TruthfulQA evaluation protocol, baseline method prompts, configuration and hyperparameters, failure cases, examples of extracted facts, generated queries, and semantic noise injection.


\begin{table}[t]
\centering
\caption{Main results on LongForm Bio and TruthfulQA. Fact Preservation Rate (in parentheses) indicates the ratio of facts retained compared to the Initial Answer. SERC demonstrates superior preservation (e.g., 106.5\% for 14B), indicating it corrects errors by refining information rather than deleting it.}
\label{tab:main_results}
\small 
\begin{tabular}{llcccc} 
\toprule
\textbf{Model} & \textbf{Method} & \textbf{\begin{tabular}[c]{@{}c@{}}LongForm Bio\\ (FactScore)\end{tabular}} & \textbf{\begin{tabular}[c]{@{}c@{}}Avg Facts\\ (Preserve \%)\end{tabular}} & \textbf{\begin{tabular}[c]{@{}c@{}}TruthfulQA\\ (Acc\%)\end{tabular}} & \textbf{\begin{tabular}[c]{@{}c@{}}TruthfulQA\\ (Avg Score)\end{tabular}} \\ 
\midrule
\multirow{4}{*}{\textbf{8B}}  & Initial Answer & 0.5976 & 23.86 (100.0\%) & 50.0\% & 5.45 \\
                              & CoVe \cite{dhuliawala2023cove} & 0.5551 & 17.23 (72.2\%) & 47.5\% & 5.45 \\
                              & CoVe+RAG & 0.7265 & 17.70 (74.2\%) & 75.0\% & 7.55 \\
                              & RARR \cite{gao2023rarr} & 0.7999 & 22.96 (96.2\%) & 68.5\% & 7.25\\
                              & Re-Ex \cite{kim2024reex} & 0.7636 & \textBF{23.53 (98.6\%)} & 77.5\% & 7.65 \\
                              & \textBF{SERC(ours)}  & \textBF{0.8568} & 18.53 (77.7\%) & \textBF{80.0\%} & \textBF{8.25} \\ 
\midrule
\multirow{4}{*}{\textbf{14B}} & Initial Answer & 0.4850 & 21.16 (100.0\%) & 67.5\% & 6.67 \\
                              & CoVe \cite{dhuliawala2023cove} & 0.4625 & 14.85 (70.2\%) & 60.0\% & 6.30 \\
                              & CoVe+RAG & 0.7427 & 17.54 (82.9\%) & 82.5\% & 8.47 \\
                              & RARR \cite{gao2023rarr} & 0.7647 & 21.21 (100.2\%) & 70.5\% & 7.65 \\
                              & Re-Ex \cite{kim2024reex}& 0.7239 & 21.77 (102.8\%) & 84.5\% & 8.53 \\
                              & \textBF{SERC(ours)}  & \textBF{0.8146} & \textBF{22.53 (106.5\%)} & \textBF{85.0\%} & \textBF{8.62} \\ 
\bottomrule
\end{tabular}
\end{table}

\section{Results and Analysis}
\label{sec:results}

\subsection{Main Results}
\label{sec:main_results}
The experimental results demonstrate that the proposed SERC framework achieves the highest quantitative metrics across both benchmarks and model scales. As detailed in Table~\ref{tab:main_results}, SERC significantly enhances factual precision; specifically, the Llama-3-8B model showed a $43.4\%$ improvement ($0.5976 \rightarrow 0.8568$) and the Qwen2.5-14B model exhibited a $68\%$ improvement ($0.4850 \rightarrow 0.8146$) in \textBF{FactScore}. On \textBF{TruthfulQA}, the accuracy improved by $30.0$ percentage points for the 8B model and $17.5$ percentage points for the 14B model.

\noindent\textbf{Domain Optimization and Inference-time Alignment.}
The initial FactScore of Qwen2.5-14B-Instruct ($0.4850$) was lower than that of Llama-3-8B-Instruct ($0.5976$). This disparity is attributed to the models' distinct optimization focuses: while Llama-3 is highly tuned for general English reasoning, Qwen2.5 excels in multilingual, mathematical, and coding domains. Despite this pre-training domain bias, SERC successfully bridged the gap, boosting Qwen's performance to $0.8146$. This empirically proves that SERC acts as a robust inference-time alignment layer, ensuring high factual reliability regardless of the backbone model's specific optimization.

\noindent\textbf{Resolution of Ambiguity and Granular Correction.}
A critical insight lies in the analysis of Fact Preservation. While iterative methods like CoVe often suffer from a decrease in fact count (defensive deletion), SERC maintains a robust volume of validated facts.

Specifically, for the 8B model, while Re-Ex exhibits a higher preservation rate ($98.6\%$) than SERC ($77.7\%$), its significantly lower FactScore ($0.7636$ vs. $0.8568$) implies a failure to filter out hallucinations (Over-preservation). In contrast, SERC demonstrates \textit{selective preservation}, effectively pruning semantic noise.

Furthermore, for the 14B model, the preservation rate reaches 106.5\%. This increase does not imply hallucinated expansion, but rather the \textit{granular refinement} of vague assertions. For instance, when SERC corrects a broad claim (e.g., "He was a politician") into specific, verified roles (e.g., "He was a diplomat and a scholar"), the number of atomic facts naturally increases. This confirms that SERC functions as a high-resolution error-corrector that restores semantic fidelity.

\begin{table}[h]
\centering
\caption{Comparison of average token usage and performance efficiency on LongForm Bio (Llama-3-8B). Although SERC incurs a higher token cost than baselines, it provides a justified trade-off by achieving significantly superior factual precision.}
\label{tab:token_usage}

\setlength{\tabcolsep}{25pt} 

\small
\begin{tabular*}{\columnwidth}{@{\extracolsep{\fill}} lcc @{}}
\toprule
\textbf{Method} & \textbf{Average Token Usage} & \textbf{FactScore (8B)} \\
\midrule
CoVe+RAG             & 21,994 & 0.7265 \\
Re-Ex                & 27,352 & 0.7636 \\
RARR                 & 29,658 & 0.7999 \\
\textBF{SERC (Ours)} & \textBF{39,492} & \textBF{0.8568} \\
\bottomrule
\end{tabular*}
\end{table}

\noindent\textbf{Cost-Benefit Analysis: The Price of Reliability.}
Table~\ref{tab:token_usage} presents the computational cost analysis. While SERC incurs a higher token overhead ($39,492$) compared to baselines like RARR ($29,658$) or Re-Ex ($27,352$), this increase is a justified \textit{Reliability Cost} for deep semantic verification.

Unlike RARR, which often performs surface-level edits, SERC engages in a rigorous reconstruction process via the LDPC-inspired framework. This investment translates directly into a substantial performance gap; SERC surpasses RARR by approximately +5.7 points in FactScore (8B) and +5.0 points (14B). Consequently, SERC occupies the optimal trade-off point, delivering state-of-the-art fidelity that outweighs the moderate increase in computational resources.

\begin{table}[h]
\centering
\caption{Performance degradation when RAG is disabled. Reliance on intrinsic knowledge fails to improve factuality, leading to score regression.}
\label{tab:ablation_rag}

\setlength{\tabcolsep}{12pt} 

\small 
\begin{tabular*}{\columnwidth}{@{\extracolsep{\fill}} llccc @{}} 
\toprule
\textbf{Model} & \textbf{Metric} & \textbf{Initial Answer} & \textBF{SERC w/o RAG} & \textbf{SERC} \\ 
\midrule
\multirow{2}{*}{\textbf{8B}}  & FactScore    & 0.6825 & 0.6695 & \textBF{0.8568} \\
                              & Avg \# Facts & 25.15  & 17.40  & \textBF{18.53}  \\ 
\midrule
\multirow{2}{*}{\textbf{14B}} & FactScore    & 0.5426 & 0.5374 & \textBF{0.8146} \\
                              & Avg \# Facts & 21.75  & 23.90  & \textBF{22.53}  \\ 
\bottomrule
\end{tabular*}
\end{table}

\begin{table}[h]
\centering
\caption{Failure case study where local correction is applied without coarse-level alignment. The model attempts to patch individual facts, resulting in a contradictory \textit{Semantic Chimera.}}
\label{tab:case_study_failure}
\small
\begin{tabularx}{\textwidth}{l X}
\toprule
\textbf{Stage} & \textbf{Text content \& Analysis} \\ 
\midrule
\textbf{User Query} & ``Tell me a bio of Fernando da Costa Novaes'' \\ \midrule
\textbf{Initial Answer} & ``...a Brazilian \textBF{footballer who played as an attacking midfielder}...'' \\ \midrule
\textbf{Local Correction} & [Corrected] DoB, Identity (to Ornithologist) but \textBF{Failed} to remove football context. \\ \midrule
\textbf{Final Output} & ``...was a Brazilian \textBF{ornithologist and former attacking midfielder}...'' \\
\bottomrule
\end{tabularx}
\end{table}

\subsection{Ablation Study}
\label{sec:ablation}
To validate our structural rationale, we conduct an ablation study on 20 random LongForm Bio instances. As shown in Tables 3--5, removing RAG or coarse-grained alignment causes significant score regression and semantic inconsistencies. Conversely, our low-density verification maintains high precision while being 45.6\% more cost-efficient than exhaustive baselines, confirming the synergy of each component in the SERC framework.

\noindent\textbf{RAG module.}
To demonstrate the necessity of external knowledge, we deactivated the RAG module (Table~\ref{tab:ablation_rag}). The results indicate that correction attempts relying solely on intrinsic knowledge lead to quality degradation; for the 8B model, \textbf{FactScore} dropped from $0.6825$ to $0.6695$. This suggests that SLMs cannot overcome \textbf{self-bias} without a reliable external reference. Thus, RAG is a prerequisite for the SERC framework.

\noindent\textbf{Coarse-grained alignment.}
As illustrated in Table~\ref{tab:case_study_failure}, omitting coarse alignment results in a \textit{semantic chimera}. While local facts (e.g., birth dates) were corrected, the incorrect context (e.g., football career) persisted. This confirms that local ``bit flips'' cannot restore the message if the initial code deviates too far from the truth manifold. coarse-grained alignment serves as an essential fail-safe to force the global context onto the correct trajectory.

\begin{table}[t]
\centering
\caption{Efficiency and Performance Comparison. \textbf{Initial} represents the raw LLM output before correction. \textbf{SERC} significantly reduces token costs while achieving higher FactScore than the 1:1 verification baseline (High-Density).}
\label{tab:token_cost}

\setlength{\tabcolsep}{8pt} 

\small 
\begin{tabular*}{\columnwidth}{@{\extracolsep{\fill}} llcccc @{}}
\toprule
\multirow{2.5}{*}{\textbf{Model}} & \multirow{2.5}{*}{\textbf{Method}} & 
\multicolumn{2}{c}{\textbf{Efficiency}} & 
\multicolumn{2}{c}{\textbf{Performance}} \\
\cmidrule(lr){3-4} \cmidrule(lr){5-6}
& & \textbf{Tokens} & \textbf{Reduct.} & \textbf{FactScore} & \textbf{Facts (Pres.\%)} \\
\midrule

\multirow{3}{*}{\textbf{8B}} 
& Initial      & - & - & 0.6572 & 23.55 (100.0\%) \\
& High-Density & 72,634 & - & 0.7002 & 14.50 { }(61.6\%) \\
& \textBF{SERC} & \textBF{39,492} & \textBF{45.6\%} & \textBF{0.8568} & \textBF{18.53 { }(77.7\%)} \\

\addlinespace[0.6em] 

\multirow{3}{*}{\textbf{14B}} 
& Initial      & - & - & 0.5050 & 21.25 (100.0\%) \\
& High-Density & 54,407 & - & 0.8128 & 24.75 (116.5\%) \\
& \textBF{SERC} & \textBF{36,533} & \textBF{32.9\%} & \textBF{0.8146} & \textBF{22.53 (106.5\%)} \\

\bottomrule
\end{tabular*}
\vspace{0.3em}
\scriptsize
\textbf{Note.} Reduct.: Cost Reduction (\%). Pres.\%: Preservation Rate relative to the initial answer.
\end{table}

\noindent\textbf{Low-Density Verification.}
We compare SERC's sparse verification with a \textit{High-Density} (1:1) baseline that verifies each atomic fact (Table~\ref{tab:token_cost}). Although dense verification can increase coverage, it substantially increases prompt length and token usage due to repeated retrieval and redundant reasoning over overlapping fact clusters. In practice, High-Density consumed 72{,}634 tokens (8B) and 54{,}407 tokens (14B), whereas SERC reduced the cost to 39{,}492 ($-45.6\%$) and 36{,}533 ($-32.9\%$) tokens, yielding a more practical performance--cost trade-off.

Moreover, dense verification may introduce excessive evidence and correction signals, increasing over-correction risk. For 8B, Dense yields only a modest FactScore gain (0.6572$\rightarrow$0.7002) while sharply reducing Avg Facts (23.55$\rightarrow$14.50), suggesting defensive deletion. While Dense improves 14B more strongly (0.5052$\rightarrow$
0.8128; Avg Facts 21.25$\rightarrow$24.75), the cost remains disproportionately high and the process becomes overly fragmented. By grouping related facts and enforcing key constraints via sparse checks, SERC reduces redundant signals and promotes more stable refinement.

\subsection{Qualitative Analysis: Logic Propagation}
\begin{table}[h]
\centering
\caption{Example of Logic Propagation. SERC updates the dependent context (``fast bowling'' to ``ball-carrying'') after correcting the root entity type (``cricket'' to ``rugby''), ensuring global semantic coherence.}
\label{tab:logic_propagation}
\vspace{-0.5em}
\small
\begin{tabularx}{\textwidth}{l X}
\toprule
\textbf{Stage} & \textbf{Content \& Semantic Analysis}  \\ 
\midrule
\textbf{Initial Answer} & ``Josh Mansour is a \textBF{cricket player}... known for his \textBF{fast bowling skills}.''  \\ \midrule
\textbf{Syndrome Det.} & \textbf{Fact:} ``Cricket player'' $\rightarrow$ \textBF{CONTRADICTED} (Evidence: ``Rugby player''). \\ \midrule
\textbf{BP Correction} & \texttt{<fixed>}Rugby league player\texttt{</fixed>} $\rightarrow$ \textBF{Triggers update:} \texttt{<fixed>}Strong ball-carrying skills\texttt{</fixed>}. \\ \midrule
\textbf{Final Output} & ``Josh Mansour is a \textBF{rugby league player}... known for his \textBF{strong ball-carrying skills}.''  \\
\bottomrule
\end{tabularx}
\vspace{-1em}
\end{table}

To further investigate the internal mechanisms of SERC, we perform a qualitative analysis of its logic propagation capability(full traces in Supp. Sec. 3). Unlike baseline methods that often perform isolated local corrections , SERC ensures global semantic consistency by approximating the belief propagation (BP) algorithm.

As shown in Table~\ref{tab:logic_propagation}, when a root entity is corrected (e.g., from a ``cricket player'' to a rugby player), SERC recognizes that dependent attributes—such as specific skills or team associations—must also be updated to maintain logical coherence. Without this mechanism, the model produces a ``Semantic Chimera,'' where factually correct updates (e.g., identity) coexist with hallucinated contexts (e.g., ``fast bowling'' skills remaining for a rugby player). This result confirms that SERC’s information-theoretic approach successfully restores the intended information manifold from noisy semantic observations.

\section{Conclusion}
This study proposes SERC (semantic error-reduction and correction), a novel self-correction framework inspired by Error Correcting Codes (ECC), to address the critical challenge of hallucinations in large language models (LLMs). By reconceptualizing text generation as a semantic noisy channel, we treat generated responses as corrupted codewords and systematically perform error detection and restoration based on information-theoretic principles. Experimental results show that SERC achieves state-of-the-art performance across benchmarks, marking a substantial improvement in factual precision compared to strong retrieval-augmented baselines. Notably, in an 8B-scale small language model (SLM) setting, SERC attains higher factual precision than a 14B model equipped with CoVe+RAG, demonstrating robust error correction that is not dependent on parameter scale. In addition, SERC exhibits a high information preservation rate by replacing incorrect information with verified facts rather than simply deleting it, thereby offering practical value for high-fidelity generation. 

The academic contributions of this work are twofold. First, we establish a new theoretical foundation by interpreting the deep learning-based generation process as a semantic noisy channel model. By transplanting LDPC mechanisms such as sparse parity-check matrices and syndrome decoding into the text domain, we provide a principled basis for systematic hallucination mitigation. Second, we implement a model agnostic, training free methodology applicable at the prompt level, empirically showing that reliable response generation is achievable even in resource-constrained SLM environments. Future work will extend this framework to multimodal channels and optimize the verification density for real-time applications.

\section{Limitations and Ethical Considerations}
\label{sec:limitations}

\noindent\textbf{Operational Cost, Latency, and Environmental Impact.} 
While achieving a 45.6\% token reduction over exhaustive methods, SERC's multi-step pipeline incurs higher token overhead and inference latency than vanilla generation (Table~\ref{tab:token_usage}). This increased computational and environmental cost makes SERC currently better suited for offline rather than real-time applications. To mitigate latency without compromising precision, future work will investigate \textit{Check Node parallelization} within the Tanner Graph, evaluating sparse parity-checks independently.

\noindent\textbf{Dependency on External Knowledge and Bias.} 
SERC's reliability heavily depends on the RAG module's knowledge base, risking the propagation of biased or incorrect retrievals. Furthermore, reliance on Western-centric benchmarks (e.g., LongForm Bio, TruthfulQA) leaves performance in culturally diverse or non-English settings unverified, potentially leading to demographic disparities.

\noindent\textbf{Domain, Modality Constraints, and Generalization.} 
Currently optimized for entity-centric natural language QA, SERC requires adaptations for structured domains like mathematics or code. However, abstracting "atomic facts" into broader \textit{"logical semantic units"} enables extending SERC's information-theoretic framework to symbolic reasoning. Applying this semantic channel model to such domains and cross-modal environments remains a critical future direction.

\noindent\textbf{Balance between Precision and Verbosity.} 
While high preservation rates (e.g., 106.5\%) demonstrate effective granular correction, they risk verbosity. Information-theoretically, an ideal decoder reconstructs messages without extraneous details. Future iterations will incorporate length constraints to strictly align the corrected response's information density with the original codeword.

\bibliographystyle{splncs04}
\bibliography{custom}

@article{lewis2020retrieval,
  title={Retrieval-augmented generation for knowledge-intensive nlp tasks},
  author={Lewis, Patrick and Perez, Ethan and Piktus, Aleksandra and Petroni, Fabio and Karpukhin, Vladimir and Goyal, Naman and K{\"u}ttler, Heinrich and Lewis, Mike and Yih, Wen-tau and Rockt{\"a}schel, Tim and others},
  journal={Advances in neural information processing systems},
  volume={33},
  pages={9459--9474},
  year={2020}
}

@inproceedings{asai2024selfrag,
  title={Self-{RAG}: Learning to Retrieve, Generate, and Critique through Self-Reflection},
  author={Asai, Akari and Wu, Zeqiu and Wang, Yizhong and Sil, Avirup and Hajishirzi, Hannaneh},
  booktitle={International Conference on Learning Representations (ICLR)},
  year={2024}
}

@inproceedings{dhuliawala2023cove,
  title={Chain-of-Verification Reduces Hallucination in Large Language Models},
  author={Dhuliawala, Shehzaad and Komeili, Mojtaba and Xu, Jing and Raileanu, Roberta and Li, Xian and Celikyilmaz, Asli and Weston, Jason},
  booktitle={arXiv preprint arXiv:2309.11495},
  year={2023}
}

@article{ji2023survey,
  title={Survey of Hallucination in Natural Language Generation},
  author={Ji, Ziwei and Lee, Nayeon and Frieske, Rita and Yu, Tiezheng and Su, Dan and Xu, Yan and Ishii, Etsuko and Bang, Ye Jin and Madotto, Andrea and Fung, Pascale},
  journal={ACM Computing Surveys},
  volume={55},
  number={12},
  pages={1--38},
  year={2023}
}

@inproceedings{huang2024large,
  title={Large Language Models Cannot Self-Correct Reasoning Yet},
  author={Huang, Jie and Chen, Xinyun and Mishra, Swaroop and Zheng, Huaixiu Steven and Yu, Adams Wei and Song, Xinying and Zhou, Denny},
  booktitle={International Conference on Learning Representations (ICLR)},
  year={2024}
}

@inproceedings{yoran2024robust,
  title={Making Retrieval-Augmented Language Models Robust to Irrelevant Context},
  author={Yoran, Ori and Wolfson, Tomer and Ram, Ori and Berant, Jonathan},
  booktitle={International Conference on Learning Representations (ICLR)},
  year={2024}
}

@inproceedings{mallen2023trust,
  title={When Not to Trust Language Models: Investigating Effectiveness of Parametric and Non-Parametric Memories},
  author={Mallen, Alex and Asai, Akari and Zhong, Victor and Das, Rajarshi and Khashabi, Daniel and Hajishirzi, Hannaneh},
  booktitle={Proceedings of the 61st Annual Meeting of the Association for Computational Linguistics (ACL)},
  pages={9802--9822},
  year={2023}
}

@inproceedings{lin2022truthfulqa,
  title={Truthful{QA}: Measuring How Models Mimic Human Falsehoods},
  author={Lin, Stephanie and Hilton, Jacob and Evans, Owain},
  booktitle={Proceedings of the 60th Annual Meeting of the Association for Computational Linguistics (ACL)},
  pages={3214--3252},
  year={2022}
}

@inproceedings{min2023factscore,
  title={FActScore: Fine-grained Atomic Evaluation of Factual Precision in Long Form Text Generation},
  author={Min, Sewon and Krishna, Kalpesh and Lyu, Xinxi and Lewis, Mike and Yih, Wen-tau and Koh, Pang Wei and Iyyer, Mohit and Zettlemoyer, Luke and Hajishirzi, Hannaneh},
  booktitle={Proceedings of the 2023 Conference on Empirical Methods in Natural Language Processing (EMNLP)},
  pages={12076--12100},
  year={2023}
}

@inproceedings{zheng2024judging,
  title={Judging {LLM}-as-a-Judge with {MT}-Bench and Chatbot Arena},
  author={Zheng, Lianmin and Chiang, Wei-Lin and Sheng, Ying and Zhuang, Siyuan and Wu, Zhanghao and Zhuang, Yonghao and Lin, Zi and Li, Zhuohan and Li, Dacheng and Xing, Eric and others},
  booktitle={Advances in Neural Information Processing Systems (NeurIPS)},
  year={2024}
}

@article{shannon1948mathematical,
  title={A Mathematical Theory of Communication},
  author={Shannon, Claude E},
  journal={The Bell System Technical Journal},
  volume={27},
  number={3},
  pages={379--423},
  year={1948}
}

@article{gallager1962ldpc,
  title={Low-Density Parity-Check Codes},
  author={Gallager, Robert G},
  journal={IRE Transactions on Information Theory},
  volume={8},
  number={1},
  pages={21--28},
  year={1962}
}

@article{tanner1981recursive,
  title={A Recursive Approach to Low Complexity Codes},
  author={Tanner, R Michael},
  journal={IEEE Transactions on Information Theory},
  volume={27},
  number={5},
  pages={533--547},
  year={1981}
}

@article{qin2021semantic,
  title={Semantic Communications: Principles and Challenges},
  author={Qin, Zhijin and Tao, Xiaoming and Liu, Jianhua and Li, Geoffrey Ye},
  journal={arXiv preprint arXiv:2201.01389},
  year={2021}
}

@inproceedings{brown2020language,
  title={Language Models are Few-Shot Learners},
  author={Brown, Tom and Mann, Benjamin and Ryder, Nick and Subbiah, Melanie and Kaplan, Jared D and Dhariwal, Prafulla and Neelakantan, Arvind and Shyam, Pranav and Sastry, Girish and Askell, Amanda and others},
  booktitle={Advances in Neural Information Processing Systems (NeurIPS)},
  volume={33},
  pages={1877--1901},
  year={2020}
}

@article{wei2022emergent,
  title={Emergent Abilities of Large Language Models},
  author={Wei, Jason and Tay, Yi and Bommasani, Rishi and Raffel, Colin and Zoph, Barret and Borgeaud, Sebastian and Yogatama, Dani and Bosma, Maarten and Zhou, Denny and Metzler, Donald and others},
  journal={Transactions on Machine Learning Research},
  year={2022}
}

@article{weidinger2021ethical,
  title={Ethical and Social Risks of Harm from Language Models},
  author={Weidinger, Laura and others},
  journal={arXiv preprint arXiv:2112.04359},
  year={2021}
}

@article{kingma2013auto,
  title={Auto-Encoding Variational Bayes},
  author={Kingma, Diederik P and Welling, Max},
  journal={arXiv preprint arXiv:1312.6114},
  year={2013}
}

@book{pearl1988probabilistic,
  title={Probabilistic Reasoning in Intelligent Systems: Networks of Plausible Inference},
  author={Pearl, Judea},
  year={1988},
  publisher={Morgan Kaufmann Publishers}
}

@article{yan2024corrective,
  title={Corrective Retrieval Augmented Generation},
  author={Yan, Shi-Qi and Gu, Jia-Chen and Zhu, Yun and Ling, Zhen-Hua},
  journal={arXiv preprint arXiv:2401.15884},
  year={2024}
}

@article{zhang2024migres,
  title={LLMs Know What They Need: Leveraging a Missing Information Guided Framework to Empower Retrieval-Augmented Generation},
  author={Wang, Keheng and others},
  journal={arXiv preprint arXiv:2404.14043},
  year={2024}
}

@inproceedings{gao2023rarr,
  title={RARR: Researching and Revising What Language Models Say, Using Language Models},
  author={Gao, Luyu and Schulman, John and Hilton, Jacob},
  booktitle={Proceedings of the 61st Annual Meeting of the Association for Computational Linguistics (Volume 1: Long Papers)},
  pages={16477--16508},
  year={2023}
}

@article{kim2024reex,
  title={{Re-Ex}: Revising after explanation reduces the factual errors in {LLM} responses},
  author={Kim, Juyeon and others},
  journal={arXiv preprint arXiv:2402.17097},
  year={2024},
  url={https://arxiv.org/abs/2402.17097}
}

@inproceedings{jeong2024adaptive,
  title={Adaptive-rag: Learning to adapt retrieval-augmented large language models through question complexity},
  author={Jeong, Soyeong and Baek, Jinheon and Cho, Sukmin and Hwang, Sung Ju and Park, Jong C},
  booktitle={Proceedings of the 2024 Conference of the North American Chapter of the Association for Computational Linguistics: Human Language Technologies (Volume 1: Long Papers)},
  pages={7036--7050},
  year={2024}
}

@article{xie2021deep,
  title={Deep learning enabled semantic communication systems},
  author={Xie, Huiqiang and Qin, Zhijin and Li, Geoffrey Ye and Juang, Biing-Hwang},
  journal={IEEE transactions on signal processing},
  volume={69},
  pages={2663--2675},
  year={2021},
  publisher={IEEE}
}

\end{document}